\documentclass{article}
\usepackage{ijcai21}

\usepackage{times}

\usepackage{soul}
\usepackage{url}
\usepackage[hidelinks]{hyperref}
\usepackage[utf8]{inputenc}
\usepackage[small]{caption}
\usepackage{graphicx}
\usepackage{amsmath}
\usepackage{amsfonts}
\usepackage{booktabs}
\usepackage[]{subfig} 
\usepackage[frozencache=true,cachedir=./minted-output]{minted}

\setmintedinline{fontsize=\footnotesize,breaklines, breakafter=_}

\newcommand{\code}[1]{\mintinline{python}{#1}}

\newif\ifdraft
\drafttrue
\usepackage{xcolor}

\def\codesize{\scriptsize} 

\title{Evaluating Continual Learning Algorithms \\ by Generating 3D Virtual Environments}

\author{
Enrico Meloni$^{1,2}$\footnote{Contact Author}\and
Alessandro Betti$^2$\and
Lapo Faggi$^{1,2}$\and
Simone Marullo$^{1,2}$\and \\
Matteo Tiezzi$^1$\And
Stefano Melacci$^1$\\
\affiliations
$^1$DINFO, University of Florence\qquad
$^2$DIISM, University of Siena\\
\emails
{\small meloni@diism.unisi.it, alessandro.betti2@unisi.it,
 \{lapo.faggi,simone.marullo\}@unifi.it,
 \{mtiezzi,mela\}@diism.unisi.it}
\vskip -3pc
}

\begin{document}

\maketitle

\begin{abstract}
Continual learning refers to the ability of humans and animals to incrementally learn over time in a given environment. 
Trying to simulate this learning process in machines
is a challenging task, also due to the inherent difficulty in creating conditions for designing continuously evolving dynamics that are typical of the real-world. Many existing research works usually involve training and testing of virtual agents on datasets of static images or short videos, considering sequences of distinct learning tasks. However, in order to devise continual learning algorithms that operate in more realistic conditions, it is fundamental to gain access to rich, fully-customizable and controlled experimental playgrounds. Focussing on the specific case of vision, we thus propose to leverage recent advances in 3D virtual environments 
in order to approach the automatic generation of potentially life-long dynamic scenes with photo-realistic appearance. Scenes are composed of objects that move along variable routes with different and fully customizable timings, and randomness can also be included in their evolution.
A novel element of this paper is that scenes are described in a parametric way, thus allowing the user to fully
control the visual complexity of the input stream the agent perceives. These general principles are concretely implemented exploiting a recently published
3D virtual environment. The user can generate scenes without the need of having strong skills in computer graphics, since all the generation facilities are exposed through a simple high-level Python interface. We publicly share the proposed generator.
\end{abstract}

\section{Introduction}
{\let\thefootnote\relax\footnotetext{Accepted for publication at the International Workshop on Continual Semi-Supervised Learning (CSSL) at IJCAI 2021 (DOI: TBA)}}
Traditional machine learning techniques usually assume static input data and the existence of a neat distinction between a training and a test phase. Input data, entirely available at the beginning of the learning procedure, are processed as a whole, iterating over the training dataset multiple times,
optimizing the performance with respect to a given learning task. The trained models are then freezed and exploited for inference only, hence computationally expensive re-training procedures are needed to possibly incorporate any new available information. This learning paradigm is clearly incompatible with what humans (and, more in general, animals) do in their everyday life, continuously acquiring and adapting their knowledge to the dynamic environment in which they live. The field of machine learning that aims at simulating this learning process by an artificial agent is known as continual or life-long learning \cite{parisi2019continual,van2019three}. The agent should be enough malleable to integrate new knowledge and, at the same time, enough stable to retain old information (\textit{stability-plasticity dilemma} \cite{abraham2005memory}).  Vanilla neural networks have been shown to struggle in this aspect, since training a network to solve a new task will likely override the information stored in its weights (catastrophic forgetting \cite{mccloskey1989catastrophic,french1999catastrophic}). In the context of computer vision (specifically, object recognition), continual learning algorithms are trained and their perfomance assessed on datasets containing static images (such as MNIST \cite{mnist} or Caltech-UCSD Birds-200 \cite{wah2011caltech}) or short sequences of temporally coherent frames (e.g. CORe50 \cite{lomonaco2017core50}), usually considering a sequence of distinct learning tasks. 
However,
the resulting learning scenarios are still far away from the original idea of an agent learning from a continuous stream of data in a real-world environment (see also the \textit{task-free} continual learning approach of \cite{aljundi2019task}). Furthermore, having the possibility to fully control the visual scene the agent perceives (number and types of objects that are present, their pose and their motion, background, possible occlusions, lighting, etc.) is essential to devise a suitable and feasible continual learning protocol and, from this point of view, real-world footages are not a viable alternative. 

We thus propose to exploit the recent technological advancements in 3D virtual environments to parametrically generate photo-realistic scenes in a fully controlled setting, easily creating customizable conditions for developing and studying continual learning agents. 
As a matter of fact, in the last few years,
due to the improved quality of the rendered scenes, 3D virtual environments have been increasingly exploited by the machine learning community for different research tasks \cite{beattie2016deepmind,gan2020threedworld,kolve2017ai2,savva2019habitat,weihs2020allenact,xia2020interactive} and different environments, based on different game engines, have been proposed so far, such as DeepMind Lab \cite{beattie2016deepmind} (Quake III Arena engine), VR Kitchen \cite{gao2019vrkitchen}, CARLA \cite{dosovitskiy2017carla} (Unreal Engine 4), AI2Thor~\cite{kolve2017ai2}, CHALET~\cite{yan2018chalet}, VirtualHome~\cite{puig2018virtualhome}, ThreeDWorld~\cite{gan2020threedworld}, SAILenv~\cite{DBLP:conf/icpr/MeloniPTGM20} (Unity3D game engine), HabitatSim~\cite{savva2019habitat}, iGibson~\cite{xia2020interactive}, SAPIEN~\cite{xiang2020sapien} (other engines). 
Moreover, the recent work~\cite{lomonaco2020continual} proposed a novel non-stationary 3D benchmark based on the VIZDoom environment to tackle model-free continual reinforcement learning.

Motivated by this significant amount of research activities, we propose to exploit such technologies to implement a method for the  generation of synthetic scenes with different levels of complexity, and that depends on well-defined customizable parameters. Each scene includes dynamical elements that can be subject to random changes, making the environment a continuous source of potentially new information for continual learning algorithms. Another key aspect in the context of continual learning is related to the source of supervisions. 3D environments can naturally provide full-frame labeling for the whole stream, since the identity of the involved 3D objects is known in advance. This paves the way to the customization of active learning technologies, in which the agent asks for supervision at a certain time and coordinates, that the 3D environment can easily provide. Moreover, in the context of semi-supervised learning, it is of course straightforward to instantiate experimental conditions in which, for example, supervisions are only available during the early stages of life of the agent, while the agent is asked to adapt itself in an unsupervised manner when moving towards a new scene. On the other hand, one could also devise methods where the learning model evolves in an unsupervised manner and the interactions with the supervisor only happen at later stages of development (i.e., for evaluating the developed features).  Finally, we introduce the perspective in which scenes could be just part of the same ``big'' 3D world, and the agent could move from one to another without abrupt interruptions of the input signal.

This paper is organized as follows.
In Section \ref{sec:generation}, the proposed generative framework is described, where
possible factors of variations will be encoded parametrically. 
Section \ref{sec:sailenv} will present a practical implementation of these ideas extending a recent 3D virtual environment, SAILenv~\cite{DBLP:conf/icpr/MeloniPTGM20}. Some illustrative examples will be given in Section \ref{sec:examples}. Finally, Section \ref{sec.conclusions} will draw some conclusions.
\section{Parametric Generation of Environments}\label{sec:generation}
This work focusses on the problem of generating customized 3D visual environments to create  experimental conditions well suited for
learning machines in a \emph{continual learning} scenario. 
In this section we describe the conceptual framework that allows us to 
formally introduce the automatic generation of a family of dynamic visual scenes. 
One of the main strengths of the automatic generation of 3D environments is the possibility to easily change and adapt them to facilitate the creation of benchmarks with 
different \textit{degrees of difficulty} with respect to a given 
model and task, allowing researchers to craft ad-hoc experiments to evaluate specific skills of the continual learning model under study or to design a range of gradually harder learning problems.

The three key factors that we consider in order to devise an automatic generator of dynamic 3D scenes are  
visual quality, reproducibility and user-control in the generation procedure.
First of all, it is important that the visual quality of the rendered scene is good enough to simulate photo-realistic conditions. On the other hand, a flexible generator should not be constrained to such high-level quality and should be able to handle also more elementary scenes in which, for instance, objects are geometric primitives or they have no or poor textures. 
At the same time, the generating
procedure should be easy to reproduce. The dynamics of the scene should be controllable at the point in which it is possible to go back to the very beginning of the agent life to reproduce the exact same visual stream; of course, this does not exclude pseudo-randomic behaviour of the environment as, in that case, the reproducibility can be guaranteed by explicitly fixing the initial condition
of the driving pseudo-random process (seed).
Scenes with high visual quality and reproducible conditions can readily be obtained as soon as one relies, for the visual
definition 
of the scenes, on a modern graphical engine which is capable of physics simulations, as we will show in our actual implementation in Section~\ref{sec:sailenv}.

Concerning the capability of customizing the generated scenes, the quality of the generator depends on the flexibility it offers in terms of compositional properties and user accessibility to such properties.
To this aim, we parametrically describe the visual 
world assuming that we have at our disposal a collection of pre-designed visual scenes $S=\{s_1,\dots, s_n\}$. For each scene $s_j$, a definite collection of object templates $\Omega_j=\{\omega_{1,j},\dots \omega_{n_j, j}\}$ is available, 
where $n_j$ is the number of object templates in the $j$-th scene. 
Each $s_j$ is initially populated by some static instances of the object templates. The parametric generation procedure instantiates new objects from the template list, eventually including multiple instances of the same template (e.g., positioning them in different locations of the 3D space--for example, a table with \textit{four} chairs). 
Formally, 
fixing a scene $\sigma\equiv s_j\in S$ with templates $\Omega \equiv \Omega_j$,
we can define the collection of $N$
objects that will be added to $\sigma$ by the parametric generation procedure as $\Phi:=(\varphi_1,\dots,\varphi_N)\in\Omega^N$.
For example, given a scene with templates $\Omega = \{\mathtt{chair}, \mathtt{pillow}, \mathtt{laptop}\}$, we could  have $\Phi = (\mathtt{chair 1}, \mathtt{chair2}, \mathtt{pillow1}, \mathtt{laptop 1})$, where $N = 4$ and we used numerical suffixes to differentiate repeated instances of the same object template.

In this work, we assume 
that the lighting conditions of the rendering engine are fixed and so the position and the 
orientation of the agent point of view.\footnote{Here we are making this assumption
in order to simplify the management of the generation procedure, however these settings can be 
regarded as additional parameters that can be chosen to 
define the environment.} 
We denote with $(v_k)_{k\in\mathbb N}$ the sequence of frames captured by the agent point of view. 
Hence, $\sigma$ can be generated once $\Phi$ is chosen and the following attributes are specified for each $\varphi_i$:
\begin{itemize}
\item the indices $(k_i,\hat k_i)\in\mathbb N^2$ of the frames where $\varphi_i$ makes respectively 
its first and last appearance;
\item the position and the orientation of the object in the frame $k_i$, collectively represented as a vector\footnote{Again, for the sake of simplicity,
we are assuming to 
work with objects which are rigid bodies (hence the 
$\mathbb R^6$) but indeed this is by no means
a crucial assumption.}
$\pi_i\in\mathbb R^6$;
\item its trajectory (i.e., its position and orientation) for each $k$ such that
$k_i < k \leq \hat k_i$, modeled by a set of parameters indicated with $\tau_i$ and defined in what follows.
\end{itemize}
Notice that, in order to grant additional flexibility to the scenario definition, it is useful to allow the possibility of dynamically spawning new objects on the fly, when the agent is already living in the generated environment. This property enables the creation of scenes that might also significantly change over time, being expanded or connected to other scenes, capabilities that might be very appropriate in the context of continual learning. The values of $(k_i,\hat k_i)$, $\pi_i$, and $\tau_i$, for $i=1,\dots, N$ are regarded 
as parameters that characterize the customizable objects visible in a frame $k$.
In particular, parameters $\tau_i$, $i=1,\dots,N$ unambiguously define the object trajectories, such as
the trajectory's global shape, the speed and whether or not  the 
trajectory completely lies in the agent's field of view. Formally, considering the $i$-th object, we have that $\tau_i = (\kappa_i,\vartheta^1_i,\dots,\vartheta^m_i)$, where $\kappa_i$ specifies the chosen kind of trajectory while $\vartheta^1_i,\dots,\vartheta^m_i$ stand for all the additional parameters required to fully determine it. Overall, the visual environment is specified by the 
collection of parameters $\Theta:=(k_1,\dots, k_N,\hat k_1,\dots,\hat k_N,\pi_1,
\dots,\pi_N,\tau_1,\dots,\tau_N)$. 

Hence it is clear that through 
the choice of $\Theta$ we can control the number of objects present at any given frame $k$,
the position and orientations of the objects,
the way in which objects moves and their velocity, i.e., the nature of their trajectories and whether or not objects escape the field of view.
A fine control over this set of parameters provide us with a general tool to create highly customizable  datasets suitable for continual learning scenarios, possibly of increasing difficulty with respect to a given learning task. For example, in an object recognition problem, the number of angles from which an object is seen, which is closely related to the chosen trajectory, could clearly affect the visual complexity of the task.

\begin{figure*}
\centering
\hbox to\hsize{\hfil
\frame{\includegraphics[width=0.5\columnwidth]{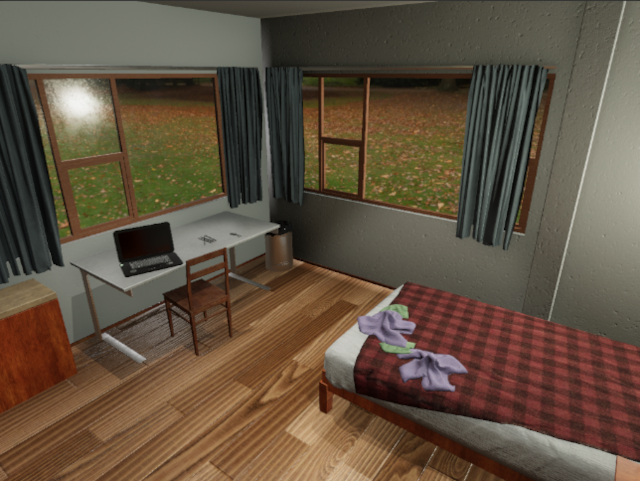}}
\hfil
\frame{\includegraphics[width=0.5\columnwidth]{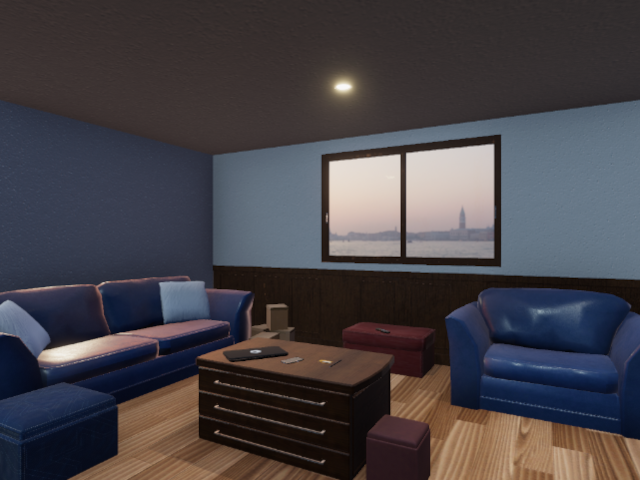}}
\hfil
\frame{\includegraphics[width=0.5\columnwidth]{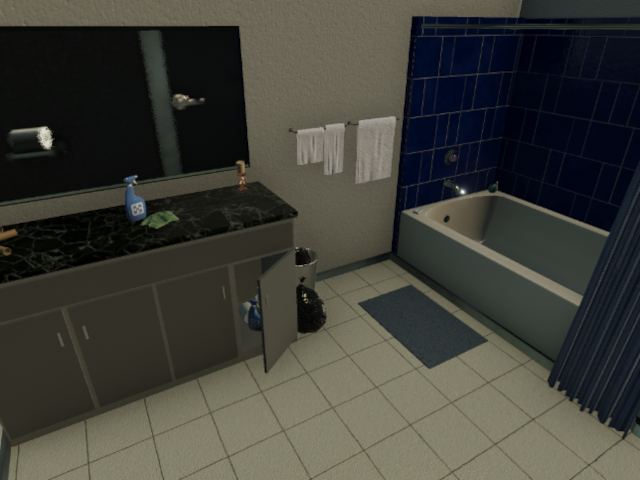}}
\hfil}
\vskip -1mm
\caption{The three default scenes of SAILenv, \texttt{room01}, \texttt{room02}, \texttt{room03} (besides the 
empty scene \texttt{object\_view}).}
\label{fig:scenes}
\vskip-1pc
\end{figure*}

\section{Continual Learning 3D Virtual Benchmark} \label{sec:sailenv}
SAILenv \cite{DBLP:conf/icpr/MeloniPTGM20} is a platform specifically designed to ease the creation of customizable 3D environments and their interface with user-defined procedures. With a few lines of code, any learning algorithm can get several data from the virtual world, such as pixel-level annotations.
SAILenv includes a Unity library with ready-to-go 3D objects and it provides basic tools to allow the customization of a virtual world within the Unity 3D editor, without the need of writing 3D graphics specific code. Differently from the other existing solutions, it also offers motion information for each pixel of the rendered view. 
SAILenv is based on the Unity Engine\footnote{See \url{https://unity.com}}, a state-of-the-art graphics and physics engine that is commonly used for videogames and physical simulations. It therefore presents realistic objects and scenes, with fine details and realistic illumination, while allowing the creation of credible motion dynamics of objects in the scene.
The SAILenv platform, when executed, creates the virtual world, managing the physical simulation in all its aspects. It also opens up a network connection listener, which waits for incoming connections to interact with the environment. The communication is implemented with low-level socket operations and a custom protocol which focuses on achieving high performance, avoiding bottlenecks in data exchange that would excessively slow down every simulation, for reasons not-related to machine learning. 

The platform is released with a Python API, which offers a high-level interface, called \textit{Agent}, that acts as the main player in the communication between the 3D world and custom Python code. The API allows the creation of multiple agents that ``live'' in the virtual world, each of them with its own view of the environment. Each agent is defined by several parameters, such as the resolution of the rendered image that is acquired from the 3D scene, its position and orientation. 
By means of a few lines of code, an agent can return fully-annotated views of the environment:
\vskip-10pt
\begin{minted}[frame=lines,bgcolor=white,fontsize=\codesize]{python}
from sailenv.agent import Agent

agent = Agent(width=256, height=192, 
              host="192.168.1.3", port=8085)
agent.register()
agent.change_scene(agent.scenes[2])

while True:
    frame_views = agent.get_frame()
    ...
agent.delete()
\end{minted}
\vskip-10pt

The data (\code{frame_views})
provided by the agent include: the \textit{RGB View} (pixel colors of the rendered scene), \textit{Optical Flow} (motion)\footnote{A pixel of the Optical Flow View is a vector $(v_x, v_y) \in \mathbb{R}^2$ representing the velocity in $\mathtt{px}/\mathtt{frame}$. For visualization purposes (e.g. see the Optical Flow rows of Figures \ref{fig:no_trail_empty}, \ref{fig:no_trail_room}, \ref{fig:trail_room_2}), each vector could be converted in polar coordinates $(\rho, \phi)$ and the pixel could be assigned the HSV color $(\phi, 1, \rho)$. Therefore, $\rho$ would determine the intensity of the color (the faster, the brighter), while $\phi$ would determine the color (red: left, green: down, cyan: right, violet: up).}, \textit{Semantic Segmentation} (category-level labels), \textit{Instance Segmentation} (instance-level labels), and \textit{Depth View} (depth). Each of these elements contains pixel-wise dense annotations. They are all generated in real-time, and they are then transmitted to the Python client with a fast low-level communication mechanism. 
This facilitates the use of the SAILenv platform in real-time online learning scenarios. 

For the purpose of this work, we extended the SAILenv platform to support dynamic scene generation following the guidelines of Section~\ref{sec:generation}. The new Python API we developed also allows the customization of the scene without having to deal with 3D-graphics editing tools or the Unity Editor, creating new objects on-demand.

\paragraph{Scenes and objects.}
We extended the SAILenv Python API to allow an easy and quick definition of the parameters in $\Theta$, through few lines of code. After having registered the Agent in the environment (as shown in the previous code snippet), a pre-designed scene $\sigma$ can be chosen using the method \code{agent.change_scene(scene_name)}. In particular, SAILenv comes with the following scenes,   $S=\{$\code{object_view} (empty space), \code{room01} (bedroom), \code{room02} (living room), \code{room03} (bathroom)$\}$ 
(see Figure \ref{fig:scenes}). Selecting a scene automatically determines the set $\Omega$ of available templates. Given a certain template, a new object $\varphi_i$ can be generated through the method 
\code{agent.spawn_object(template_name, position, rotation[, dynamic, limited_to_view])}, specifying its position, rotation and, in the case of a moving object, the properties of the  associated trajectory (last two arguments).  
This method will return an \code{object_id}.
Invoking the creation at frame $k$ will spawn the selected object at the next frame ($k_i=k+1$) and it will set $\pi_i$ to the concatenation
of the given position and rotation.
We postpone the description of the trajectory dynamics (\code{dynamic} argument) to the next paragraph, while when the Boolean flag \code{limited_to_view} is set to true, the object will be always kept withing the the field of view of the agent. The condition for making this choice effective is to create invisible barriers  where the object will bounce, located at the borders of the agent camera frustum (that is the region of 3D world seen by the agent), by calling \code{agent.spawn_collidable_view_frustum()}.
The object can then be deleted through the method \code{agent.despawn_object(object_id)} which is equivalent to setting $\hat{k}_i$ to the identifier of the next frame.

\paragraph{Trajectories.}
The object dynamics can be defined through simple Python classes. In this work, we propose three different types of trajectories, associated to classes that can be instantiated by calling: \code{LinearWaypoints(waypoints_list, total_time)}, \code{CatmullWaypoints(waypoints_list, total_time)} and \code{UniformMovementRandomBounce(speed, angular_speed, start_direction[, seed])}. Within the notation of Section 2, the chosen class trajectory for the $i$-th object is what we formalized with $\kappa_i$ (for example,  consider $\kappa_i$ set to \code{UniformMovementRandomBounce}), while the associated arguments (\code{speed, angular_speed, start_direction[, seed]}, in the case of the previous example) stand for $\vartheta^1_i,\dots,\vartheta^4_i$. 
Both \code{CatmullWaypoints} and \code{LinearWaypoints} require a list of $L$ waypoints $(w_1,\dots, w_L) \in (\mathbb R^6)^L$ and the time (in seconds) that the object takes to loop around all of them, see Figure \ref{fig:empty_scenario} for an example of code (described in the next section). The difference between the two dynamics is that the former does a linear interpolation between two consecutive waypoints, while the latter computes a Catmull-Rom Spline interpolation \cite{maggini2007representation} along the whole set of  waypoints. Collisions with other scene elements are handled by the Unity physics engine, that takes care of rejoining the trajectory whenever it becomes possible.
\code{UniformMovementRandomBounce} makes an object move inertially until it hits another one or the edges of the agent view. After the collision, the object bounces back in a random direction and  also acquires an additional random torque. The \code{speed} and the \code{angular_speed} parameters limit the velocity of the object in the scene, the \code{start_direction} bootstraps its movement and the \code{seed} may be fixed to replicate the same dynamics (i.e., for reproducibility purposes). Furthermore, the API allows to change the object position and orientation at any given time through the method \code{agent.move_object(object_id, position, rotation)}. 

\paragraph{Utilities.} What we described so far fully defines the scene and the parameters in $\Theta$. 
In order to simplify the management of the Python code, 
we added a higher abstraction level based on the Python class \code{Scenario} and some additional utility classes, such as \code{Waypoint} and \code{Object}, that allow to describe the structure of the scene in a compact manner, as we will show in the examples of Section~\ref{sec:examples} (Figure \ref{fig:empty_scenario}, \ref{fig:room_scenario} and \ref{fig:room_01_scenario}). When using class Scenario, the object trajectories can be orchestrated through the \code{Timings} classes.
There are three different  available timings. The first one, \code{AllTogether(wait_time)}, makes every object  move at the same time after \code{wait_time} seconds. The second, \code{WaitUntilComplete},  supports only waypoint-based dynamics (more, generally, dynamics that are based on loops), and activates them one at a time waiting until each one is complete before starting the next one. Finally, the  \code{DictTimings(_map)} timing takes as input a map that defines for each trajectory how long it should be active before stopping and starting the next one. Finally, we mention the \code{Frustum} class to simplify the creation of the previously described invisible boundaries, if needed.

\begin{figure}[t]
    \centering
\begin{minted}[frame=lines,bgcolor=white,fontsize=\codesize]{python}
scene = "object_view/scene"
waypoints = [
    Waypoint(Vector3(0., 0., 4.), Vector3(0., 0., 0.)),
    ...
    Waypoint(Vector3(-5., 1., 7.), Vector3(90., 90., 180.))
]
dynamic = CatmullWaypoints(waypoints=waypoints, total_time=10.0)
objects = [
    Object("c1", "Cylinder", 
           Vector3(0, 0, 2), Vector3(0, 0, 0), dynamic)
]
scenario = Scenario(scene, objects)
agent.load_scenario(scenario)
\end{minted}
\vskip -5mm
    \caption{
    A Cylinder moves through the defined waypoints, with a trajectory obtained by Catmull interpolation.}
    \label{fig:empty_scenario}
\vskip-5pt
\begin{minted}[frame=lines,bgcolor=white,fontsize=\codesize]{python}
scene = "room_02/scene" 
dynamic1 = UniformMovementRandomBounce(seed=32, 
                   speed=0.8, start_direction=Vector3(0, 5, 2))
dynamic2 = UniformMovementRandomBounce(...) 
dynamic3 = UniformMovementRandomBounce(...) 
agent_pos = agent.get_position()
objects = [
    Object("c1", "Chair 01",  agent_pos + Vector3(2, 0, 0), 
           Vector3(0, 0, 0),  dynamic1, frustum_limited=True),
    Object("p1", "Pillow 01", ...), 
    Object("d1", "Dish 01", ...)
] 
timings = AllTogetherTimings(0.75)
view_limits = Frustum(True, 10.)
scenario = Scenario(scene, objects, timings, view_limits)
agent.load_scenario(scenario)
\end{minted}
\vskip -5mm
    \caption{Definition of a simple scenario where a Chair, a Pillow and a Dish move pseudo-randomly around a pre-built living room.}
    \label{fig:room_scenario}
\vskip-1pc
\end{figure}

\section{Examples}
\label{sec:examples}
\begin{figure}[!hb]
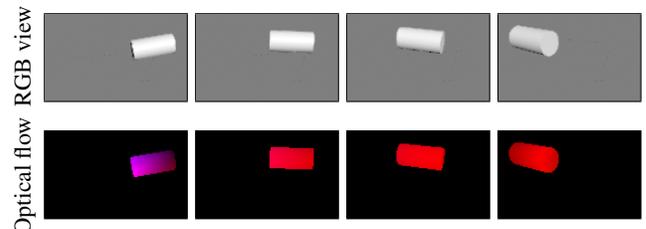


\def\|#1|{$\vcenter{\hbox{\frame{\includegraphics[width=0.22\columnwidth,trim={2cm 2.5cm 2.5cm 1.5cm},clip]{figures/empty/no_trail/#1.png}}}}$}
\tabskip=1em plus2em minus.7em
\halign to\hsize{#\hfil &#&#&#&#\cr
\rotatebox[origin=c]{90}{\small RGB view}&\|216_main|&\|224_main|&\|232_main|&\|240_main|\cr
\noalign{\smallskip}
\rotatebox[origin=c]{90}{\small Optical flow}&\|216_flow|&\|224_flow|&\|232_flow|&\|240_flow|\cr
}
\vskip -3mm
\caption{\label{fig:no_trail_empty} Scene described by the script in Figure \ref{fig:empty_scenario}. Four frames are shown (from left to right)---RGB view and 
optical flow.}
\end{figure}

The proposed SAILenv-based generator can be downloaded at SAILenv official website \url{https://sailab.diism.unisi.it/sailenv/}. In the following we show three examples of generations.
\paragraph{Example 1.} In Figure~\ref{fig:empty_scenario} the SAILenv basic scene named \code{object_view} is chosen, that is an empty space with monochrome background. Then, a set of waypoints is defined and the dynamic \code{CatmullWaypoints} is created using them. A single object is specified, named \code{c1}, based on template \code{Cylinder}, at position \code{Vector3(0,0,2)} and with an initial orientation 
specified by \code{Vector3(0,0,0)} (Euler angles);
here \code{Vector3(_,_,_)} is the description of a 
three dimensional vector. The \code{CatmullWaypoints} dynamics will move the Cylinder through each waypoint, interpolating the trajectory with a Catmull-Rom spline. Using the notation presented in Section \ref{sec:generation}, we have: 
$\sigma = \mathtt{object\_view}$, $\Omega = \{\ldots, \mathtt{Cylinder}, \dots\}$, $\Phi = (\mathtt{c1})$, $(k_1, \hat{k}_1) = (0, \infty)$
and the associated trajectory is specified by $\kappa_1 =$ \code{CatmullWaypoints}, $\vartheta^1_1 = \mathtt{waypoints}$ and $\vartheta^2_1 = \mathtt{total\_time} =10$.
The generated RGB view and the corresponding optical flow are shown in Figure \ref{fig:no_trail_empty} considering four different time instants.



\begin{figure}[t]
    \centering
\begin{minted}[frame=lines,bgcolor=white,fontsize=\codesize]{python}
scene = "room_01/scene" 
waypoints = [ 
    Waypoint(Vector3(0.5, 1.4, 0.5), Vector3(0., 0., 0.)),
    Waypoint(Vector3(0.3, 1., -1.), Vector3(90., 0., 0.)),
    ... 
]
agent_pos = Vector3(-1.3, 2., 1.5)
agent.set_position(agent_pos)
agent.set_rotation(Vector3(22., 144., 0))
dynamic = CatmullWaypoints(waypoints=waypoints)
objects = [
    Object("racket", "Tennis Racket 01", 
            Vector3(0.5, 1.4, 0.5), Vector3(0., 0., 0.), dynamic)
]
scenario = Scenario(scene, objects)
agent.load_scenario(scenario)
\end{minted}
    \vskip -5mm
\caption{A Tennis Racket moves along a set of waypoints 
(Catmull interpolation) inside a pre-built bedroom.}
    \label{fig:room_01_scenario}
\vskip-1pc
\end{figure}

\begin{figure*}
\def\|#1|{$\vcenter{\hbox{\frame{\includegraphics[width=0.49\columnwidth]{figures/room/full3/#1.png}}}}$}
\tabskip=1em plus2em minus.7em
\halign to\hsize{#\hfil &#&#&#&#\cr
\rotatebox[origin=c]{90}{\small RGB view}&\|100_main|&\|106_main|&\|112_main|&\|118_main|\cr
\noalign{\smallskip}
\rotatebox[origin=c]{90}{\small Semantic segmentation}&\|100_cat|&\|106_cat|&\|112_cat|&\|118_cat|\cr
\noalign{\smallskip}
\rotatebox[origin=c]{90}{\small Optical flow}&\|100_flow|&\|106_flow|&\|112_flow|&\|118_flow|\cr
}
\caption{\label{fig:no_trail_room} Scene described by the script in Figure \ref{fig:room_scenario} (\code{room02}---livingroom scene) considering four different frames (from left to right). For each object, the chosen dynamic is \code{UniformMovementRandomBounce}. For each frame we display the RGB view, the semantic segmentation and the 
optical flow. Additionally, we depict in the RGB and semantic segmentation views the local trajectories followed by the moving objects (attached to the moving objects).}

\bigskip
\def\|#1|{$\vcenter{\hbox{\frame{\includegraphics[width=0.49\columnwidth]{figures/room2/trail/#1.png}}}}$}
\tabskip=1em plus2em minus.7em
\halign to\hsize{#\hfil &#&#&#&#\cr
\rotatebox[origin=c]{90}{\small RGB view}&\|060_main|&\|070_main|&\|080_main|&\|090_main|\cr
\noalign{\smallskip}
\rotatebox[origin=c]{90}{\small Semantic segmentation}&\|060_cat|&\|070_cat|&\|080_cat|&\|090_cat|\cr
\noalign{\smallskip}
\rotatebox[origin=c]{90}{\small Optical flow}&\|060_flow|&\|070_flow|&\|080_flow|&\|090_flow|\cr
}
\caption{\label{fig:trail_room_2} 
Scene described by the script in Figure  \ref{fig:room_01_scenario} (\code{room01}---bedroom scene) considering four different frames (from left to right). The chosen dynamic is \code{CatmullWaypoints}.
For each frame we display the RGB view, the semantic segmentation and the 
optical flow. Additionally, we depict in the RGB and semantic segmentation views the full trajectory followed by the moving object (attached to the racket).}
\end{figure*}


\paragraph{Example 2.} In Figure \ref{fig:room_scenario} the selected pre-designed scene is \code{room_02}, a realistic living room with common furniture. The novel SAILenv API allows us to add new objects that, in this case, are a \code{chair}, a \code{pillow} and a \code{dish}, from the templates \code{Chair 01}, \code{Pillow 01} and \code{Dish 01}. They are initially located in specific
points relative to the agent's position (\code{agent_pos + Vector3(_,_,_)}) with a certain orientation (the second \code{Vector3(_,_,_)}). For all the objects, the dynamic \code{UniformMovementRandomBounce} is chosen, also specifying their speed, their initial direction and the seed to ensure the reproducibility of the pseudo-random bounces.
Finally, the \code{AllTogetherTimings} configuration is selected, making every object move at the same time within the view frustum of the agent and also never going beyond 10 meters of distance from the agent itself (\code{view_limits=Frustum(True,10.)}). Using the notation of Section \ref{sec:generation}, we have 
$\sigma = \mathtt{room\_02}$, $\Omega = \{$\ldots,\code{Chair 01}, \ldots,\code{Pillow 01}, \ldots, \code{Dish 01}$, \dots\}$, $\Phi = ($\code{c1}, \code{p1}, \code{d1}$)$ and $(k_i, \hat{k}_i) = (0, +\infty)$ $\forall\,\, i$. Moreover, $\kappa_i =$\code{UniformMovementRandomBounce} with possible different $\vartheta_i^m$ (\code{seed}, \code{speed}, \code{start_direction}) $\forall\,\, i$. 
For an illustration of the final result, see Figure \ref{fig:no_trail_room} (RGB view, semantic segmentation and optical flow). 

\paragraph{Example 3.} Finally, the code in Figure \ref{fig:room_01_scenario} illustrates another realistic scene (bedroom) in which a tennis racket moves according to the \code{CatmullWaypoints} dynamic. The selected waypoints are defined at the beginning of the script, toghether with the initial position and orientation of the racket. According to the notation of Section \ref{sec:generation},  we have $\sigma =\hbox{\code{room_01}}$, $\Phi = ($\code{racket}$)$ from the template \code{Tennis Racket 01} and $(k_i, \hat{k}_i) = (0, +\infty)$ $\forall i$. In this last case, $\kappa_i =$\code{CatmullWaypoints} and $\vartheta_1^1$ = \code{waypoints}. The final result is shown in Figure \ref{fig:trail_room_2}. Notice that we also used SAILenv facilities to change the position and orientation of the agent.

\section{Conclusions}\label{sec.conclusions}
In this paper we have proposed the idea of generating fully customizable 
datasets to train and test continual learning agents through the use of 3D-virtual environments. Describing the generating process of the scenes parametrically allows the user to have full control on the final visual stream the agent perceives and, given a certain learning task, to create scenarios of increasing difficulty. We have reported a concrete realization of these ideas in the SAILenv virtual environment, showing the potential effectiveness of this approach.  

\subsubsection{Acknowledgements}
This work was partially supported by the PRIN 2017 project RexLearn (Reliable and Explainable Adversarial Machine Learning), funded by the Italian Ministry of Education, University and Research (grant no. 2017TWNMH2).

\clearpage
\bibliographystyle{named}
\bibliography{bib}

\end{document}